\documentclass[10pt,twocolumn,letterpaper]{article}

\usepackage{iccv}
\usepackage{times}
\usepackage{epsfig}
\usepackage{graphicx}
\usepackage{amsmath}
\usepackage{amssymb}
\usepackage{authblk}

\usepackage[inline]{enumitem}
\usepackage{booktabs}
\usepackage{multirow}
\usepackage{tabularx}
\newcolumntype{C}{>{\centering\arraybackslash}X}
\usepackage[accsupp]{axessibility}  

\setlist{itemsep=0.1cm,topsep=0pt,parsep=0pt,partopsep=0pt,leftmargin=0.3cm}  
\usepackage{xcolor}
\usepackage{comment}
\usepackage{caption}

\usepackage[pagebackref=true,breaklinks=true,letterpaper=true,colorlinks,bookmarks=false]{hyperref}

\usepackage[capitalize]{cleveref} 
\crefname{section}{Sec.}{Secs.}
\Crefname{section}{Section}{Sections}
\Crefname{table}{Table}{Tables}
\crefname{table}{Tab.}{Tabs.}
\Crefname{equation}{Equation}{Equations}
\crefname{equation}{eq.}{eqs.}
\crefformat{equation}{Eq.~#2#1#3} 

\input{dc_macros}

\newcommand{\R}{\mathbb{R}}
\newcommand*{\indicator}{\text{\usefont{U}{bbold}{m}{n}1}}

\newcommand{\SAP}{\widetilde{\textrm{AP}}}
\newcommand{\VSAP}{\overrightarrow{\textrm{AP}}}

\newcommand\blfootnote[1]{%
  \begingroup
  \renewcommand\thefootnote{}\footnote{#1}%
  \addtocounter{footnote}{-1}%
  \endgroup
}

\iccvfinalcopy 

\def\iccvPaperID{10721} 

\ificcvfinal\fi

\begin{document}

\title{LoCUS: Learning Multiscale 3D-consistent Features from Posed Images}


  
  


\author{Dominik A. Kloepfer$^1$, Dylan Campbell$^2$, Jo\~ao F. Henriques$^1$\\
        $^1$Visual Geometry Group, University of Oxford\\
        $^2$Australian National University\\
        \texttt{\{dominik, joao\}@robots.ox.ac.uk, dylan.campbell@anu.edu.au}}
\maketitle
\ificcvfinal\thispagestyle{empty}\fi

\begin{abstract}
An important challenge for autonomous agents such as robots is to maintain a spatially and temporally consistent model of the world.
It must be maintained through occlusions, previously-unseen views, and long time horizons (\eg, loop closure and re-identification).
It is still an open question how to train such a versatile neural representation without supervision.
We start from the idea that the training objective can be framed as a \emph{patch retrieval} problem: given an image patch in one view of a scene, we would like to retrieve (with high precision and recall) all patches in other views that map to the same real-world location.
One drawback is that this objective does not promote \emph{reusability }of features: by being unique to a scene (achieving perfect precision/recall), a representation will not be useful in the context of other scenes.
We find that it is possible to balance retrieval and reusability by constructing the retrieval set carefully, leaving out patches that map to far-away locations.
Similarly, we can easily regulate the scale of the learned features (\eg, points, objects, or rooms) by adjusting the spatial tolerance for considering a retrieval to be positive.
We optimize for (smooth) Average Precision (AP), in a single unified ranking-based objective.
This objective also doubles as a criterion for choosing landmarks or keypoints, as patches with high AP.
We show results creating sparse, multi-scale, semantic spatial maps composed of highly identifiable landmarks, with applications in landmark retrieval, localization, semantic segmentation and instance segmentation. 
\blfootnote{Code and model weights for this project can be found at \url{https://www.robots.ox.ac.uk/~vgg/research/locus}.}
\end{abstract}


\begin{figure}[!t]\centering
   \includegraphics[width=\linewidth]{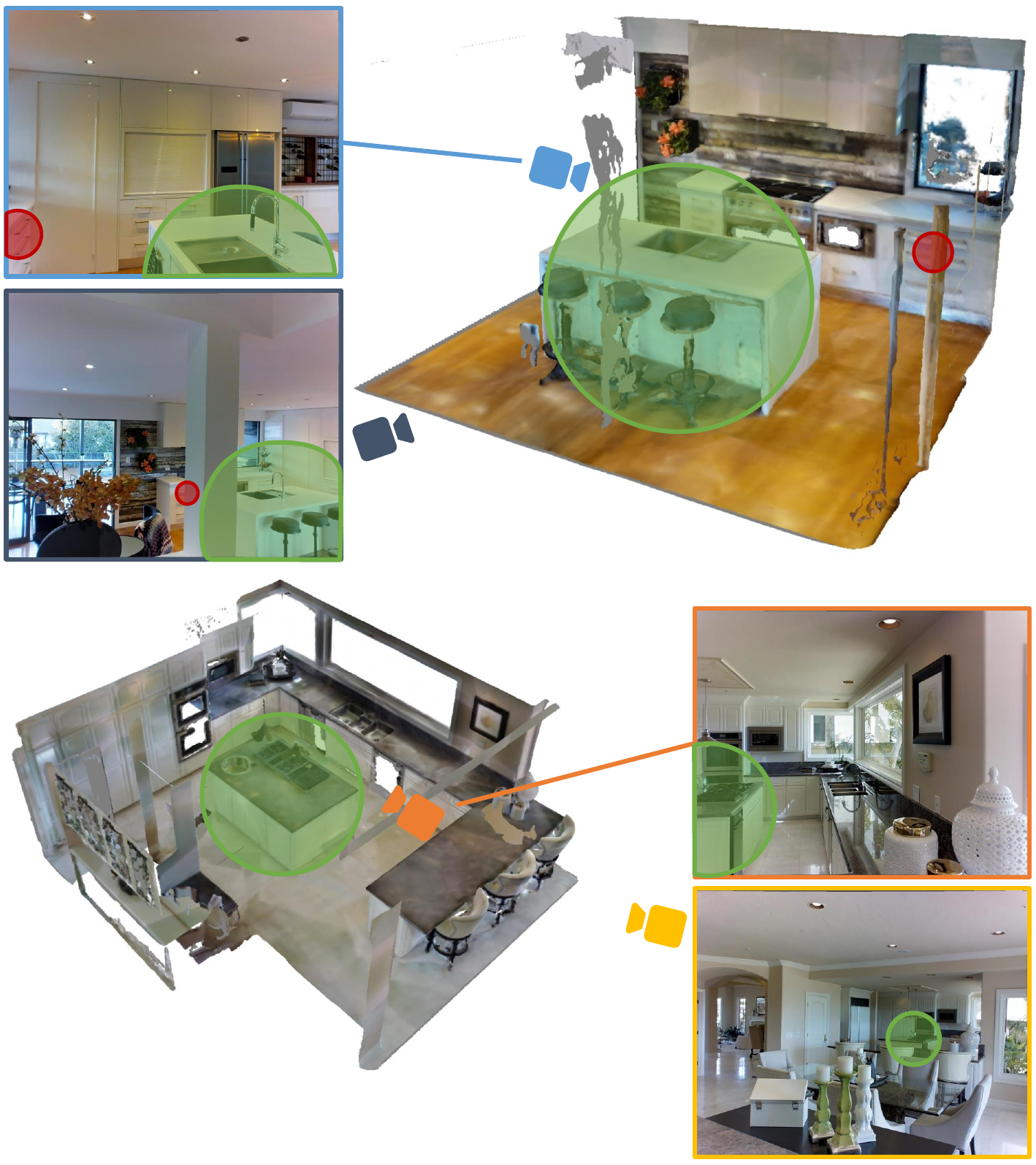}
   \vspace{-21pt}
   \caption{
   Problem setting.
   Our goal is to train a network to extract features that are identifiable and 3D-consistent, so that features at image locations corresponding to the same region in 3D space, but viewed from different positions, are similar. This can be done at multiple scales, from large (\eg, the kitchen islands in the large green circles) to small (\eg, the drawers in the small red circles).
   However, simply optimizing for ``unique'' representations at each location (\eg, via contrastive learning) runs the risk of over-fitting to the training scenes, as such objectives will discourage reuse of the same representation for different places.
   Instead, we encourage reusable landmark representations, such as the concept of a kitchen island, which may appear in different scenes (top and bottom panels) with appearance variations. 
   }
   \label{fig:splash}
\end{figure}

\section{Introduction}
\label{sec:intro}

For an autonomous agent to be able to take useful actions, it must maintain a spatially and temporally consistent world model.
This model may comprise the agent itself (\eg, pose estimation), the environment (\eg, mapping), and dynamic objects (\eg, instance detection), and must be maintained when confronted with previously-unseen views, occlusions, and long time horizons (\eg, loop closure and re-identification).
However, visual observations used by an agent for this purpose are inherently inconsistent: the same landmark may appear significantly different from different viewpoints in space and time, due to (self) occlusion, reflections, lighting variations, and dynamic effects, among other factors.
Errors in the estimation of the agent's internal state compound these problems.
Therefore, it is important for any vision-based agent to convert observations into some spatio-temporally consistent form.

Existing approaches \cite{brahmbhatt2018geometry, Henriques18a, Tschernezki22} do this at the observation synthesis (mapping) stage, by aggregating or distilling visual information in 3D. 
We argue that significant progress can be made before this point, at the observation processing stage, which lends itself to a more flexible image-centered representation that is useful for a range of tasks.
The key is to encourage consistency between image features that unproject to the same region of 3D space, within a spatial tolerance, defining a landmark at a given scale.

This can be achieved by formulating the problem as one of patch retrieval: given an image patch from one view of a scene, retrieve all patches in other views that correspond to the same 3D location, with high precision and recall.
To encourage reusability, so that the learned features are useful in new scenes, we exclude patches from the retrieval set if (when unprojected) they exceed a fixed distance from the query.
Excluding such patches ensures that the representations of similar-looking landmarks in distant places are not pushed apart unnecessarily, which would promote over-fitting unique representations to the training scenes, thus making them non-reusable in new scenes.
Moreover, by adjusting the spatial tolerance that defines the positive set, we can regulate the scale of the learned features.
This allows us to learn features at a small scale (\eg, local textures and structures), medium scale (\eg, household objects), and large scale (\eg, whole rooms or places) in the same framework.
\\

We learn this representation by optimizing a ranking-based metric, (smooth) Average Precision (AP), which doubles as a criterion for choosing distinctive landmarks (keypoints).
The resulting Location-Consistent Universal Stable (LoCUS) features are semantically-meaningful, 3D-consistent at the selected scale, and balance distinctiveness with reusability, producing sparse, multi-scale, and semantic maps.
We demonstrate applications in landmark retrieval, localization, semantic segmentation and instance segmentation.
\\

To summarize, our contributions are:
\begin{enumerate}
    \item A framework for learning 3D-consistent features from posed images via retrieval, taking into account multiple scales and how to trade off retrieval performance vs. generalization performance (reusability).
    \item A unified ranking-based objective function that facilitates the selection of highly-identifiable landmarks.
    \item An evaluation of the proposed features on real images of indoor environments, on the tasks of place recognition, semantic segmentation, instance segmentation and re-identification, as well as relative pose estimation.
\end{enumerate}

\section{Related work}
\label{sec:related_work}

The topics of keypoint detection and description~\cite{detone2018superpoint, dusmanu2019d2, revaud2019r2d2}, feature matching~\cite{yi2018learning, dang2018eigendecomposition, fathy2018hierarchical, sarlin2020superglue, sun2021loftr}, structure-from-motion \cite{schonberger2016structure}, and SLAM \cite{engel2014lsd, mur2015orb, teed2021droid} have a rich history. 
Here, we concentrate on the most recent and related work. 

\paragraph{Image retrieval.}
Learning representations for image retrieval---the task of ranking all instances in a retrieval set according to their relevance to a query image---is well-studied \cite{arandjelovic2016netvlad, radenovic2016cnn, gordo2016deep}. Metric learning approaches use, \eg, contrastive \cite{chopra2005learning} or triplet \cite{weinberger2006distance} losses to encourage positive instances to be close, while negative instances are separated by a margin. Other approaches optimize (approximations to) ranking-based metrics like Average Precision (AP) directly \cite{rolinek2020optimizing, Brown20}. For example, Smooth-AP~\cite{Brown20} proposes a sigmoid relaxation of the ranking function, where the tightness of the approximation is controlled by the temperature. Optimizing a ranking metric allows a model to target the correct ranking without caring about the absolute feature distances.
We leverage the image retrieval literature by defining our learning task as a patch retrieval problem. By carefully defining the retrieval set, we can balance feature distinctiveness with re-usability. While image retrieval methods retrieve entire images, we retrieve 3D spherical regions projected to 2D. That is, while methods such as Brown \etal~\cite{Brown20} compute a single feature per image that is then used to retrieve other images of the same class, we compute features for pixel patches that are then used to retrieve pixel patches that cover (parts of) the same 3D spherical region. More details can be found in \cref{sec:method}.

\paragraph{Learning visual features and keypoints.}
Several works explore methods to learn better image features or keypoints to facilitate 2D--2D matching \cite{yi2018learning, dang2018eigendecomposition, fathy2018hierarchical, sarlin2020superglue, sun2021loftr} or 2D--3D matching \cite{dang2018eigendecomposition, campbell2020solving} for relative/absolute pose estimation or triangulation. For example, Fathy \etal \cite{fathy2018hierarchical} use metric learning to learn 2D--2D matchable features, while Campbell \etal \cite{campbell2020solving} learn geometric features that facilitate 2D--3D matching via an end-to-end trainable blind PnP solver.

Keypoint detectors, by contrast, aim to find a sparse set of repeatable points in an image~\cite{detone2018superpoint, dusmanu2019d2, revaud2019r2d2}. For example, SuperPoint~\cite{detone2018superpoint} jointly computes keypoints and descriptors using a convolutional network trained in a self-supervised framework. Similarly, D2-Net~\cite{dusmanu2019d2} obtains keypoints via non-maximum suppression on the learned feature maps. R2D2~\cite{revaud2019r2d2} argues that repeatable regions are not necessarily discriminative, so learns to predict keypoint repeatability and reliability separately.
Unlike the features learned in these works, the features that optimize our loss function do not vary as rapidly, allowing them to more closely resemble the real scene geometry and enabling segmentation.

\paragraph{Neural mapping and reconstruction.}
Deep learning approaches have gradually closed the gap on classical Structure-from-Motion \cite{schonberger2016structure} and SLAM \cite{engel2014lsd, mur2015orb} approaches to mapping and reconstruction. For example, Neural Radiance Fields (NeRF) \cite{mildenhall2020nerf} has demonstrated photorealistic reconstruction for known cameras, and been extended to RGBD SLAM \cite{sucar2021imap, zhu2022nice}, RGB SLAM \cite{zhu2023nicer}, and semantic mapping \cite{zhi2022ilabel}.
Earlier, MapNet~\cite{Henriques18a} investigated neural localization and mapping through convolution operators, resulting in an environment map that stores multi-task information distilled from the RGBD input, which exhibits emergent semantic meaning.
Our approach produces very different kinds of maps: sparse, multi-scale, and semantic, composed of highly identifiable landmarks.

\paragraph{Self-supervised visual feature learning.}
Vision transformers (ViT) \cite{dosovitskiy2021image} have demonstrated a strong capacity for learning useful and meaningful features from large amounts of unlabelled image data~\cite{caron2021emerging, hamilton2022unsupervised, wang2023cut}. For example, DINO~\cite{caron2021emerging} demonstrated that self-supervised ViT features could be used for unsupervised object segmentation. The model was trained via self-distillation between a student network and a momentum teacher network that receive two different random transformation of an image and are encouraged to encode similar features.

STEGO~\cite{hamilton2022unsupervised} extends DINO to unsupervised semantic segmentation via contrastive learning. It trains a shallow segmentation network appended to a fixed DINO backbone with contrastive terms that encourage the learned features to form compact clusters while preserving their global relationships.
CutLER~\cite{wang2023cut} extends DINO to unsupervised object detection and segmentation, achieving extremely compelling results. The model generates training data for a detector by creating foreground object masks using normalized cuts on the patch-wise similarity matrix of DINO features, with additional object masks being found through an iterative masking procedure.

N3F~\cite{Tschernezki22} showed that DINO image features can be distilled into a 3D feature field using the same rendering loss as NeRF~\cite{mildenhall2020nerf}, given camera pose supervision. They demonstrate that the resulting features are 3D-consistent, enabling 3D instance segmentation and scene editing.
Our approach builds on these self-supervised methods by proposing a proxy patch retrieval task defined in 3D, unlike STEGO and CutLER, allowing us to adapt DINO features so that they learn invariances to viewing direction and instance. Like N3F, we require camera pose supervision to enable our 3D-aware loss. Unlike N3F, our features are defined in image space and can be predicted from a single image, facilitating applications like relative pose estimation.

\section{Method}
\label{sec:method}

Our training procedure will be centered on the concept of recognizing \emph{landmarks}: regions of space that are visually identifiable and unique within a bounded region, but reusable outside that region. We mean that landmark embeddings (representations) are ``reusable'' in the sense that the same embedding may be shared by more than one landmark, as long as they are far away in the spatial domain.

Assume that we are given a set of training images, divided into $n$ (potentially overlapping) rectangular patches $x_{i}$, \ie, the receptive fields of a Convolutional Neural Network (CNN) or the tokens of a Visual Transformer (ViT). Each training patch $x_{i}\in\cP$ is also associated with an environment $e_{i}\in\cE$ (e.g. the identity of a house in a training set composed of distinct houses) and real-world coordinates within that environment $p_{i}\in\R^{3}$, obtained for example by projecting the center coordinates of the patch using known camera geometry (camera pose and approximate depth) \cite{Hartley04c}. Note that this information is only needed for training -- at test time no such information is necessary. The training set is then $\cX=\{(x_{1},e_{1},p_{1}),\ldots,(x_{n},e_{n},p_{n})\}$.

Assume that we have also defined a set of \emph{tentative landmarks} $\cL=\{(\theta_{1},\epsilon_{1},\ell_{1}),\ldots,(\theta_{m},\epsilon_{m},\ell_{m})\}$ in 3D space: points $\ell_{i}\in\R^{3}$ in environments $\epsilon_{j}\in\mathcal{E}$ and associated embeddings $\theta_{i}\in\R^{c}$. These do not have to correspond to actual landmarks (or identifiable locations in 3D), and can be sampled uniformly across space.\footnote{We will discuss more efficient sampling strategies in Section \ref{sec:sampling}.}

We wish to train a deep neural network $\phi:\mathcal{P}\mapsto\R^{c}$ to output embeddings that can be used to match each patch $x_{i}$ to a landmark embedding $\theta_{j}$, by computing pairwise scores
\begin{equation}\label{eq:similarity}
s_{ij}=\frac{\phi(x_{i})\transpose\theta_{j}}{\|\phi(x_{i})\|\|\theta_{j}\|},
\end{equation}
consisting of a cosine distance (inner product of normalized embeddings), where higher scores denote more likely matches. To specify whether a match is correct or not, we place a sphere of radius $\rho_{j}$ around the landmark $\ell_{j}$, and any retrievals there (and in the same environment) are considered positive:
\begin{equation}
y_{ij}^{+}=\indicator\left(\left\Vert p_{i}-\ell_{j}\right\Vert \leq\rho_{j}\wedge e_{i}=\epsilon_{j}\right),\label{eq:pos-set}
\end{equation}
where $\indicator(\cdot)\in\{0,1\}$ is the indicator function. We will use $y_{ij}^{+}$ as a binary mask to denote positive matches, while $y_{ij}^{\Omega}=1$ is a trivial mask that denotes the union of positives and negatives. Both are used to define the Smooth Average Precision (Smooth-AP):
\begin{equation}
\SAP_{j}=\frac{1}{\sum_{i}^{n}y_{ij}^{+}}\sum_{i}^{n}y_{ij}^{+}\frac{1+\sum_{kl}^{nm}y_{kl}^{+}\sigma_{\tau}(s_{kl}-s_{ij})}{1+\sum_{kl}^{nm}y_{kl}^{\Omega}\sigma_{\tau}(s_{kl}-s_{ij})},\label{eq:sap}
\end{equation}
with the sigmoid $\sigma_{\tau}(x)=\frac{1}{1+\exp(-x/\tau)}$. In the limit $\tau\rightarrow0$, $\SAP_{j}$ recovers the exact AP with $\theta_{j}$ as the query embedding.

\paragraph{Discussion.}
\Cref{eq:sap} is similar to Smooth-AP, proposed by Brown \etal \cite{Brown20}, with a few differences that were necessary to adapt it to patch-based landmark retrieval: 1) the retrieval set consists of rectangular image patches, so $\phi$ can be applied convolutionally; 2) the positive set is defined by 3D Euclidean distance (\cref{eq:pos-set}) with per-landmark radii $\rho_{j}$; and 3) we wrote \cref{eq:sap} as a function of binary masks $y_{ij}^{+}$ and $y_{ij}^{\Omega}$, instead of nested sets.

This objective encourages the features from two image patches to be similar if they correspond to 3D locations that are at most a distance $\rho$ apart, since they will be in each other's positive sets. Thus the objective directly encourages 3D-location-consistent features, extracting similar features for different viewpoints of the same 3D location. Empirical support for this is given in \cref{subsec:exp_place_recognition}.
The objective also encourages semantic meaningfulness, extracting similar features for image patches that correspond to the same object.
First, note that if two 3D locations are separated by greater than $\rho$ but less than $2\rho$, they are both within the positive set of a third landmark location, encouraging all three features to be similar.
Second, note that the Smooth-AP loss does not minimise the similarity of a landmark with patches in the negative set, it only encourages the similarity with respect to the positive set to be greater than that with the negative set.
Together, this results in similar features being extracted across an object, facilitating segmentation.

\paragraph{Multi-scale landmarks.}
The radius $\rho_{j}$ of each landmark defines its overall scale, as any matching embeddings $\phi(x_{i})$ must be invariant to different positions within this radius. Thus $\phi$ may learn to recognize not only small-scale keypoints, but also landmarks at the scale of household objects, whole rooms or even larger regions (place recognition), as illustrated in fig. \ref{fig:splash}.

Despite these changes, Smooth-AP still offers a few other challenges to be adapted to our setting, which we will detail in the next sections.

\begin{figure}[t]
\begin{center}
   \includegraphics[width=\linewidth]{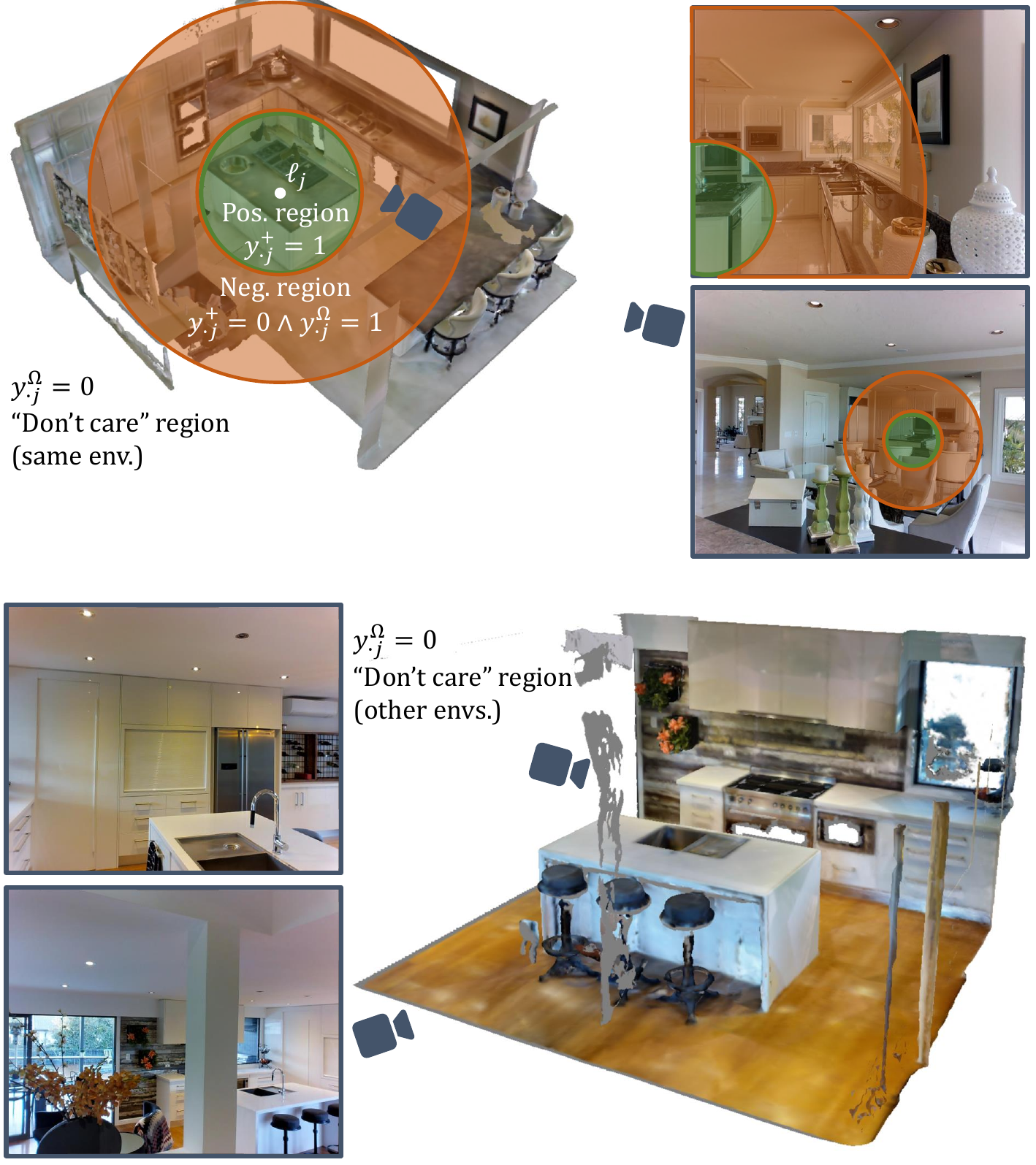}
\end{center}
   \caption{
   Illustration of the projections of the spherical regions that define the landmark retrieval objective (sec. \ref{sec:reusability}).
   The small green sphere around the tentative landmark $\ell_j$ defines the region inside which image patches are considered positive matches with the landmark ($y^+_{ij}=1$).
   The larger orange sphere defines the region with positive and negative matches ($y^\Omega_{ij}=1$).
   Importantly, outside this region matches are ignored ($y^\Omega_{ij}=0$), as well as in other environments (bottom panel).
   As a result, a contrastive (or retrieval) self-supervised objective does not suppress similar embeddings for semantically-similar but spatially distant landmarks, such as the two kitchen islands in the two environments shown.
   }
\end{figure}

\subsection{Landmark reusability: ``don't-care'' regions}\label{sec:reusability}

Optimizing for AP has one unfortunate side effect: every high score matching a patch $x_{i}$ further away from a (tentative) landmark $\ell_{j}$ than $\rho_{+}$ will be treated as a false positive, and thus suppressed during training. Likewise, all patches in different environments $e_{i}\neq\epsilon_{j}$ will be treated the same. While this seems reasonable on the surface, at the optimum it will force all landmarks to be \emph{unique} to a particular place in an environment and thus useless in a new environment or far away location. We would like some landmarks to be \emph{reusable} and shared among different environments, for example for one landmark to represent a living room in different homes, as opposed to overfitting to a single living room.

In analogy with ``don't-care'' conditions in digital circuit design \cite{micheli1994synthesis}, which reduce circuit complexity by freeing up modeling capacity for input--output combinations that are not important, we propose to define ``don't-care'' regions where the Smooth-AP objective does not constrain the deep network's output.
Instead of the trivial mask $y_{ij}^{\Omega}=1$ that denotes the universe of all patches as positives and negatives, we instead reduce this universe to
\begin{equation}
y_{ij}^{\Omega}=\indicator\left(\left\Vert p_{i}-\ell_{j}\right\Vert \leq\kappa\rho_{j}\wedge e_{i}=\epsilon_{j}\right),\label{eq:neg-set}
\end{equation}
with $\kappa>1$ a multiplier for the distance threshold. Together, $\kappa$ and $\rho_{j}$ define two concentric regions: a sphere of radius $\rho_j$ around a landmark, where any retrievals are considered positive (\cref{eq:pos-set}), and a spherical shell at distance $d$ from the landmark, with $\rho_{j}<d\leq\kappa\rho_{j}$, where any retrievals are considered negative (\cref{eq:neg-set}). Any points outside the radius $\kappa\rho_{j}$ are not considered as part of the retrieval set, and are not assigned a label. The end result is that two different tentative landmarks can have very similar embeddings $\theta_{j}$, as long as they are at a distance greater than $\kappa\rho_{j}$, and this embedding reuse will not be penalized by the Smooth-AP (\cref{eq:sap}).

\subsection{Automatic landmark selection with Vectorized-Smooth-AP}

So far we referred to landmarks as ``tentative'', so that they may not correspond to actual identifiable regions of space. However, optimizing for \cref{eq:sap} assumes a landmark $\ell_{j}$ is fixed as a query. If we maximize \cref{eq:sap} in expectation over $j$ (analogously to Brown \etal \cite{Brown20}), we implicitly give equal importance to all tentative landmarks, even if some may correspond to places that are not easily identifiable (\eg, a wall or empty region).

Rather than devise a heuristic to identify good landmarks, we instead just let the Smooth-AP objective focus on pairs of landmarks and patches that maximize AP, by considering all pairs as if they're part of a single query 
\begin{equation}
\VSAP=\frac{1}{\sum_{ij}^{nm}y_{ij}^{+}}\sum_{ij}^{nm}y_{ij}^{+}\frac{1+\sum_{kl}^{nm}y_{kl}^{+}\sigma_{\tau}(s_{kl}-s_{ij})}{1+\sum_{kl}^{nm}y_{kl}^{\Omega}\sigma_{\tau}(s_{kl}-s_{ij})}.\label{eq:vsap}
\end{equation}
\Cref{eq:vsap} is equivalent to \emph{vectorizing} the matrix of masks $Y^{+}\in\R^{n\times m}$ with elements $y_{ij}^{+}$, by stacking its elements into a single vector $\mathbf{y}^{+}\in\R^{nm}$, and computing the Smooth-AP objective (\cref{eq:sap}) with this modified input. While subtle, this has the effect that $\VSAP$ will be maximized by first distinguishing the easiest landmark--patch pairs from the rest, while ignoring those that are too ambiguous. By neglecting to emphasize all tentative landmarks equally, the objective adaptively selects highly distinguishable landmarks. We can identify them by evaluating the \emph{non-vectorized} Smooth-AP ($\SAP_{j}$) on each individually, and taking the top-$k$ landmarks:
\[
\ell^{*}=\underset{j}{\textrm{top-}k}\;\SAP_{j}.
\]

\subsection{Sampling tentative landmarks}\label{sec:sampling}

We now turn to the definition of the tentative landmark positions $\ell_{j}$ and embeddings $\theta_{j}$.

\paragraph{Sampling positions $\boldsymbol{\ell_{j}}$.}

While ideally it would be sufficient to sample the landmark positions randomly across space (either within a bounded region, or restricted to the visible hull), in a mini-batch with limited memory this is often not efficient. The reason is that $y_{ij}^{+}$ may have too few non-zero values due to non-intersecting image views, especially with a limited number of images in an environment, or in very large environments.

We found that sampling uniformly across space is very inefficient, as over $94\%$ of the chosen tentative landmarks are not visible by more than 2 views (in the training set of Matterport3D \cite{chang2017matterport3d}; see sec. \ref{sec:experiments} for details on the experimental setting).
This creates very poor query sets for retrieval, with only one or two positive embeddings, which cause over-fitting as the network easily attains $100\%$ AP on such tentative landmarks.
Instead, we need to bias the sampling more towards more visible locations.
Thankfully, there is a simple way to sample spatial positions proportionally to how often they are visible in the training set of views: simply sample uniformly among all image patches across all training images. This guarantees that the sampled distribution is proportional to how often a 3D position is visible, and is easy to implement.

\paragraph{Sampling embedddings $\boldsymbol{\theta_{j}}$.}

A straightforward way to define the embedding for $\ell_{j}$ is to average the embeddings of all patches that map to that location in space:
\[
\theta_{j}=\frac{1}{n}\sum_{i}^{n}y_{ij}^{+}\phi(x_{i}).
\]
In practice, we found that approximating this average by a single patch $\phi(x_{i})$ such that $y_{ij}^{+}>0$ (chosen at random) is sufficient, which simplifies the implementation.

\section{Experiments}
\label{sec:experiments}

In this section, we will detail our experiments, where we evaluate the ability of LoCUS features to perform place recognition and relative pose estimation, as well as evaluate its emergent semantic properties, in the form of semantic segmentation and instance segmentation with object re-identification.

\subsection{Experimental setup}%
\label{sec:setup}

\paragraph{Datasets.}
Our primary dataset for training and evaluation is the Matterport3D dataset \cite{chang2017matterport3d}, which contains a wide variety of indoor environments, captured densely with RGB and depth information. It also includes dense segmentations at the object level, which facilitate the evaluation of our model's semantic properties.

\paragraph{Training details.}
We train a 2-stage transformer \cite{dosovitskiy2021image} with 128-dimensional internal features, on top of a frozen DINO backbone \cite{caron2021emerging}. The final features extracted from image patches have 64 dimensions, and the DINO backbone computes 768-dimensional features, so we use two linear layers to map between these feature spaces, resulting in 503,232 trainable parameters. 
This model is trained by implementing the Vectorized-Smooth-AP objective from \cref{eq:vsap}.
We maximize the objective using the Adam optimizer with an initial learning rate of $10^{-4}$ and mini-batches of size 16, sampled from the Matterport3D training set \cite{chang2017matterport3d}, and train for 20 epochs.
For all experiments, we set the hyper-parameters $\tau = 0.01$ and $\rho_j = 0.2$ (in meters). 
With these settings, the model can be trained on a single NVIDIA RTX 2080Ti GPU. 

\begin{table}[t]
\centering\small
\caption{Place recognition (retrieval) results, for our LoCUS features and the DINO~\cite{caron2021emerging} baseline. We report our objective, the smooth vectorized AP ($\VSAP$), and the Average Precision (AP), which quantifies the retrieval performance. For the same features, AP, which corresponds to our objective in the limit of $\tau \rightarrow 0$, will always be higher than $\VSAP$.}
\label{table:place_recog_results}
\renewcommand*{\arraystretch}{1.1}
\begin{tabularx}{\linewidth}{@{}l C C C C@{}}
\toprule
 & \multicolumn{2}{c}{Objective ($\VSAP$)} & \multicolumn{2}{c}{Average Precision (AP)} \\
Model & Train & Val. & Train & Val. \\
\cmidrule{1-3} \cmidrule(l){4-5}
ResNet50 \cite{resnet50} & 0.11 & 0.11 & 0.11 & 0.12 \\
DINO \cite{caron2021emerging} & 0.20 & 0.20 & 0.20 & 0.20 \\
DINOv2 \cite{oquab2023dinov2} & 0.17 & 0.17 & 0.17 & 0.17 \\
LoCUS (Ours) & \textbf{0.56} & \textbf{0.54} & \textbf{0.57} & \textbf{0.55} \\ 
\bottomrule
\end{tabularx}
\end{table}

\subsection{Place recognition and retrieval}\label{subsec:exp_place_recognition}
Since our method is trained with a specific relaxation of Average Precision (AP) on retrieval-focused sets of image patches, its primary objective is most closely aligned with place recognition via retrieval.
As such, we start by evaluating its AP on unseen validation environments, which contain objects and layouts that were not seen during training. This assesses the reusability of features produced by our method.

\paragraph{Baselines.}
For this experiment, we compare with pretrained ResNet50~\cite{resnet50}, DINO~\cite{caron2021emerging}, and DINOv2~\cite{oquab2023dinov2} baselines. The features of the final layer of the ViT are reduced to 64 using PCA, the same dimension as our features (similar to Tschernezki \etal \cite{Tschernezki22}). Since our model shares almost all of its weights with the DINO baseline, this comparison well-illustrates the effect and advantages of our training method.

\paragraph{Results.}
The results for this experiment are reported in \Cref{table:place_recog_results}.
In addition to the AP on the validation set, for both our method and the DINO~\cite{caron2021emerging} baseline, we also report the AP on the training set.
As expected, our LoCUS features significantly outperform the DINO baseline, despite sharing almost all weights. 
While the retrieval performance decreases on the validation set, this decline is minimal compared to the effect of the training, demonstrating the reusability of the features in unseen environments. 


\begin{figure}[t]
\centering
\label{fig:cosegmentation}
\begin{minipage}{.22\textwidth}
  \centering
  \includegraphics[width=\linewidth]{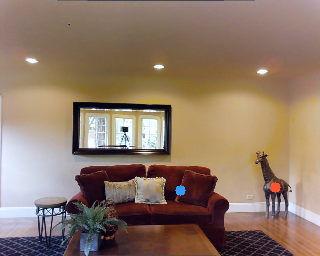}
  \label{fig:source}
\end{minipage}%
\hspace{0.5em}
\begin{minipage}{.22\textwidth}
  \centering
  \includegraphics[width=\linewidth]{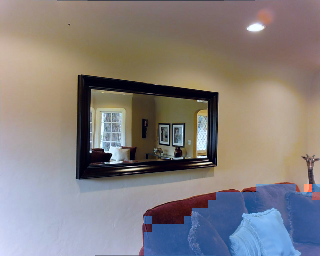}
  \label{fig:image2}
\end{minipage}\\[0.5em]
\begin{minipage}{.22\textwidth}
  \centering
  \includegraphics[width=\linewidth]{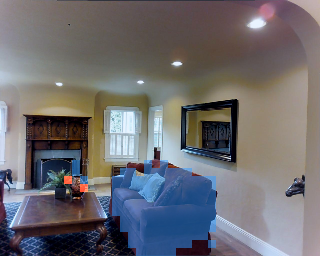}
  \label{fig:image3}
\end{minipage}%
\hspace{0.5em}
\begin{minipage}{.22\textwidth}
  \centering
  \includegraphics[width=\linewidth]{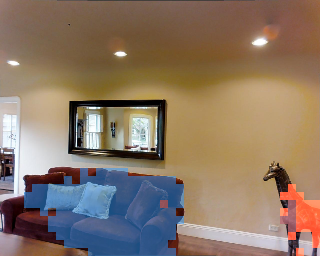}
  \label{fig:image4}
\end{minipage}
\caption{
Visualization of co-segmentation results, obtained by thresholding the cosine distance (\cref{eq:similarity}) of the LoCUS features of a query image patch (blue and orange, highlighted in the top left image) and LoCUS features of patches in other views (remaining images). The thresholded regions are indicated in matching colours.
}
\end{figure}

\begin{table}[t]
\centering\footnotesize
\caption{Semantic and instance segmentation (respectively ``stuff'' and ``things'') results, with object re-identification. Both models extract 64-dimensional feature vectors for $8\times8$ pixel patches, which are then classified into the relevant classes using a linear probe. Semantic classes contain ``stuff'' pixels grouped into their semantic categories, while instance classes contain pixels belonging to individual objects. ${}^\star$Uses ground-truth instance labels.
${}^\dagger$Released after submission deadline.}
\label{table:object_seg_results}
\renewcommand*{\arraystretch}{1.1}
\setlength\tabcolsep{3.5pt}  
\begin{tabularx}{\linewidth}{@{}l C C C C C C C C C@{}}
\toprule
 & \multicolumn{3}{c}{Semantic} & \multicolumn{3}{c}{Instance} & \multicolumn{3}{c}{Overall} \\
Model & mAP & mIoU & Jac & mAP & mIoU & Jac & mAP & mIoU & Jac \\
\cmidrule{1-4} \cmidrule(l){5-7} \cmidrule(l){8-10}
ResNet50 & 0.39 & 0.26 & 0.55 & 0.18 & 0.11 & 0.12 & 0.19 & 0.12 & 0.41 \\ 
DINO~\cite{caron2021emerging} & 0.49 & 0.34 & 0.63 & 0.40 & 0.28 & 0.28 & 0.40 & 0.29 & 0.52 \\
DINOv2${}^\dagger$ & \textbf{0.55} & \underline{0.38} & \underline{0.67} & \underline{0.49} & 0.35 & 0.39 & \underline{0.49} & 0.35 & 0.58 \\ 
Mask2Former & - & 0.03 & 0.07 & - & 0.00 & 0.00 & - & 0.00 & 0.06 \\
+ Oracle${}^\star$ & - & \textbf{0.41} & \textbf{0.71} & - & \underline{0.39} & \underline{0.53} & - & \underline{0.39} & \underline{0.64} \\
MaskDINO & - & 0.05 & 0.15 & - & 0.00 & 0.00 & - & 0.00 & 0.12 \\
+ Oracle${}^\star$ & - & \textbf{0.41} & \textbf{0.71} & - & \textbf{0.40} & \textbf{0.54} & - & \textbf{0.40} & \textbf{0.65} \\
LoCUS (Ours) & \underline{0.53} & 0.37 & \underline{0.67} & \textbf{0.54} & \textbf{0.40} & 0.42 & \textbf{0.54} & \underline{0.39} & 0.59 \\
\bottomrule
\end{tabularx}
\end{table}



\subsection{Semantic and instance segmentation}%
\label{sec:segmentation}

We now turn to scene-level object segmentation.
There are two broad categories of segmentation classes: 1)~amorphous geometry (``stuff'') such as walls, floor and ceiling; and 2)~distinct objects (``things'') such as furniture or appliances.
The former are useful for evaluating semantic segmentation at the texture level, while the latter requires distinguishing individual objects, and thus allows us to evaluate instance segmentation.
This setting is slightly more broad than instance segmentation: an object must not merely be segmented distinctly from other objects in a given image, but it must be also \emph{re-identified} in different images from varied points of view, so it also encompasses the task of object re-identification.

\paragraph{Qualitative results on co-segmentation.}
We start by exploring a single co-segmentation task, highlighting a patch in one image and then finding all matching patches in other views, by simply thresholding the similarity metric (\cref{eq:similarity}).
The results can be seen in \Cref{fig:cosegmentation}. We can observe that, despite dramatic changes in viewpoint, the learned LoCUS features are very stable over 3D space, successfully matching over very significant changes in distance, rotation, partial occlusion and out-of-view regions.

\paragraph{Implementation.}
For the quantitative evaluation, we use linear probes to assess the learned features' correlation with respect to the semantic classes, as is common in self-supervised learning \cite{caron2021emerging}.
To do this, we extract the LoCUS features $\phi(x_i)$ over all training images (considered frozen) and train a patch-wise linear classifier with a cross-entropy loss and the ground-truth segmentation labels.
We use the same optimizer settings as for the main objective until convergence, for all methods.

\paragraph{Evaluation setting and metrics.}
We use an evaluation set of \emph{unseen scenes}, which are not part of the training set, and thus test the generalization ability of the methods.
We report segmentation metrics on ``stuff'' pixels only (semantic segmentation), on ``things'' pixels only (instance segmentation and object re-identification), and on all pixels taken together.
For each case, we compute three metrics:
\begin{enumerate}
    \item mAP: For each object instance (in the case of instance segmentation) or for each class (in the case of semantic segmentation), we calculate the average precision (AP) of the linear classifier, in a one vs. all mode (\ie, considering all other pixels as negative labels). We then average across all instances or classes to obtain a mAP score.
    \item mIoU: We calculate the Intersection-over-Union (IoU) \cite{ronneberger2015unet} between the predictions and ground-truth binary masks for each object (or class) separately, and then report the average.
    \item Jaccard (Jac): Similarly to the mIoU, we compute the Jaccard index separately for each object (or class), and report the average. The Jaccard index is given by TP $/$ $($FP $+$ FN$)$, given the counts of binary True Positives (TP), False Positives (FP) and False Negatives (FN).
\end{enumerate}


\paragraph{Baselines.}
To provide a point of comparison to the semantic segmentation capabilities of the proposed features, we also report results for a number of segmentation baselines. 
We evaluate pretrained ResNet50~\cite{resnet50}, DINO~\cite{caron2021emerging}, and DINOv2~\cite{oquab2023dinov2} feature extractors, first reducing the computed features to the same number of dimensions as ours (64) using PCA computed over the full training set. This is the same strategy employed to evaluate Neural Feature Fusion Fields \cite{Tschernezki22}. Similar to our method, we then use a linear probe to produce the segmentations.

We also evaluate two recent segmentation-specific models, Mask2Former~\cite{mask2former} and MaskDINO~\cite{li2023maskdino} in their default setting, and a setting where we relabel each predicted segmentation mask with the ground-truth scene-consistent instance ID (``Oracle'').
The former performs poorly because it does not maintain consistent instance identities across frames (no object re-identification), as is required by this task.

\begin{figure}[h]
\centering
\begin{minipage}{.225\textwidth}
  \centering
  \includegraphics[width=\linewidth]{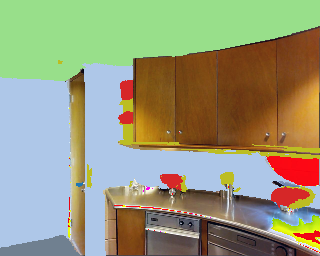}

  \label{fig:image1}
\end{minipage}%
\hspace{0.5em}
\begin{minipage}{.225\textwidth}
  \centering
  \includegraphics[width=\linewidth]{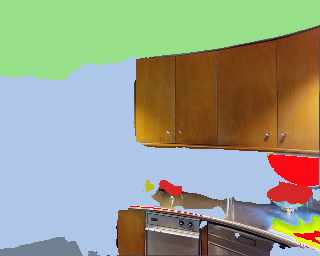}

  \label{fig:image2}
\end{minipage}\\[0.5em]

\begin{minipage}{.22\textwidth}
  \centering
  \includegraphics[width=\linewidth]{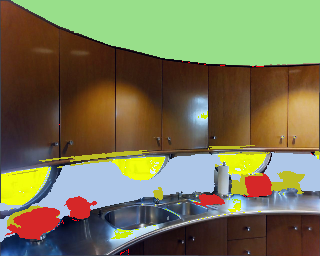}

  \label{fig:image5}
\end{minipage}%
\hspace{0.5em}
\begin{minipage}{.22\textwidth}
  \centering
  \includegraphics[width=\linewidth]{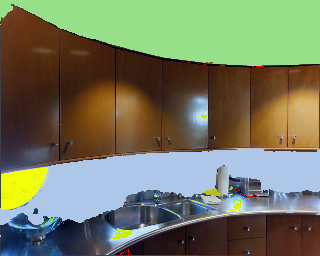}

  \label{fig:image6}
\end{minipage}\\[0.5em]
\begin{minipage}{.22\textwidth}
  \centering
  \includegraphics[width=\linewidth]{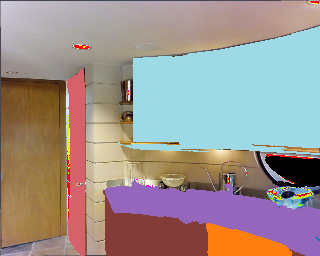}

  \label{fig:image3}
\end{minipage}%
\hspace{0.5em}
\begin{minipage}{.22\textwidth}
  \centering
  \includegraphics[width=\linewidth]{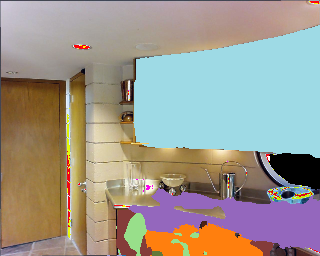}

  \label{fig:image4}
\end{minipage}\\[0.5em]

\begin{minipage}{.22\textwidth}
  \centering
  \includegraphics[width=\linewidth]{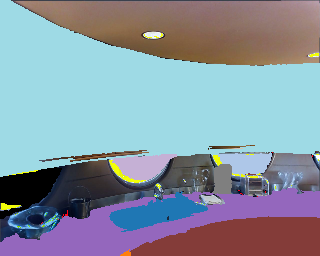}
  \captionsetup{labelformat=empty}
  \caption{Ground Truth}
  \label{fig:image7}
\end{minipage}%
\hspace{0.5em}
\begin{minipage}{.22\textwidth}
  \centering
  \includegraphics[width=\linewidth]{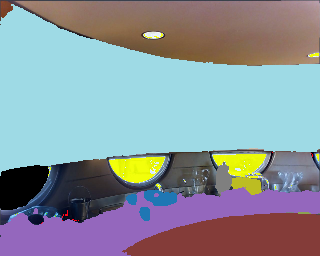}
    \captionsetup{labelformat=empty}
  \caption{Predicted}
  \label{fig:image8}
\end{minipage}
\caption{Qualitative results of semantic segmentation (wall, ceiling, floor classes) in the first two rows and of instance segmentation (household objects) in the third and fourth row. Note that object instance identities are stable across viewpoints, thus also performing object re-identification.}
\label{fig:four_rows_grid}
\end{figure}

\paragraph{Results.}
The results are shown in \Cref{table:object_seg_results}.
Our proposed LoCUS features are better able to discriminate both semantic classes, such as undifferentiated ceiling or wall regions, as well as to re-identify particular object classes.
\Cref{fig:four_rows_grid} visualizes the qualitative results. 
The proposed method outperforms the baseline feature extractor methods, especially on the instance segmentation and re-identification task, showing that the object identity predictions are stable under viewpoint changes. 
DINO features are trained to be invariant to image-space augmentations \cite{caron2021emerging}, and so understandably do not enjoy the same stability across viewpoints, especially when they change dramatically.

Our method performs comparably with the Oracle methods despite not receiving the ground-truth labels.

\subsection{Relative pose estimation}%
\label{sec:pose}

Since our LoCUS features are trained to be stable across 3D viewpoints, within a specified scale, they should be helpful for tasks that require spatial reasoning.
Furthermore, the fact that we can train features for landmarks at different scales should help with coarse-to-fine strategies.
For this reason, we focus on relative pose estimation.
Note that this is different from other settings like Simultaneous Localization And Mapping (SLAM), since those assume temporal continuity over a video stream.
In contrast, we perform relative pose estimation between single pairs of images, without any extra context, which severely limits the information available.

\paragraph{Dataset.}
We use the image pairs generated from Matterport3D for relative pose estimation that were introduced in SparsePlanes~\cite{jin2021planar}.
We remark that there is very limited overlap between the views of each pair, which makes this an extremely challenging task.

\paragraph{Metrics.}
We report standard metrics for translation and rotation error.
For translation, we report median and average errors, as well as the fraction of pairs that have an error smaller than 1 meter.
For rotation, we also report median and average errors, and the fraction of pairs with error smaller than 30 degrees.

\paragraph{Method.}
Given a pair of images, we extract the LoCUS features of each patch $\phi(x_i)$.
We then calculate pairwise scores (\cref{eq:similarity}), and for each patch, filter out all scores that are smaller than a threshold of 0.7. We then use two conditions on the continuity of the mapping from patches in image A to patches in image B to remove outliers from the set of patch pairs. Details on this process can be found in the supplementary material.
Taking the top-100 pairs by score, we use a standard robust 5-point RANSAC algorithm \cite{nister2004efficient} to calculate the essential matrix with the smallest error, and then find the corresponding relative pose (up to unknown scale) using a RANSAC chirality check \cite{bradski2000opencv}. The unknown scale can in practice be recovered using for example very coarse depth measurements; here we simply scale the translation vector by its ground truth length. 

\paragraph{Baselines.}
We compare with several baselines from the literature.
Most of these were specifically engineered for geometric matching tasks, while ours focuses on coarser (multi-scale) landmark retrieval. As such, we expect ours to be more robust at matching in the large scale, while other methods to do better at very fine-grained geometric matching.
We report results for SuperPoint \cite{detone2018superpoint} (pre-trained and with its feature dimension reduced to 64 using PCA) with nearest neighbours (NN) search and SuperPoint with FGINN for outlier removal. Given the pixel matches extracted in this way, we compute the in the same way as our method (5-point RANSAC \cite{nister2004efficient}).
We also report results for a number of methods that do not extract features, but are specialised to estimate relative poses more directly: SparsePlanes \cite{jin2021planar}, 8-Point-Supervision \cite{rockwell2022eight}, and PlaneFormers \cite{agarwala2022planeformers}. 

\paragraph{Results.}
The results are presented in \Cref{table:rel_pos_results}.
We can see that, despite not being trained specifically for camera localization, the spatial stability of the trained features does help localize the camera correctly in most instances.
Nevertheless, we would expect that with greater overlap between views, methods that are more geared towards fine-grained keypoint matching would do better than coarse matching methods such as ours, which are more concerned with coarse place (landmark) recognition. 
The most comparable method are the two relative pose estimation algorithms using SuperPoint keypoints, which our method outperforms.

\begin{table}[t]
\centering\small
\caption{Relative pose estimation results. We report translation errors in meters and rotation errors in degrees.}
\label{table:rel_pos_results}
\renewcommand*{\arraystretch}{1.1}
\setlength\tabcolsep{3pt}  
\begin{tabularx}{\linewidth}{@{}l C C C C C C@{}}
\toprule
 & \multicolumn{3}{c}{Translation} & \multicolumn{3}{c}{Rotation} \\
Model & Med. & Avg. & $\leq 1$m & Med. & Avg. & $\leq \!30^\circ$ \\
\cmidrule{1-4} \cmidrule(l){5-7}
SparsePlanes \cite{jin2021planar} & 0.63 & 1.25 & 0.67 & 7.33 & 22.78 & 0.83 \\
8-Point-Sup \cite{rockwell2022eight} & 0.64 & 1.01 & 0.67 & 8.01 & 19.1 & 0.85 \\
PlaneFormers \cite{agarwala2022planeformers} & 0.66 & 1.19 & 0.67 & 5.96 & 22.2 & 0.84 \\
\cmidrule{1-4} \cmidrule(l){5-7}
SuperPoint \cite{detone2018superpoint} & & & & & & \\
 + NN & 1.08 & 1.84 & 0.48 & 34.4 & 47.8 & 0.47 \\
 + FGINN \cite{mishkin2015mods} & 1.02 & 1.87 & 0.49 & 29.9 & 45.2 & 0.50 \\
\cmidrule{1-4} \cmidrule(l){5-7}
LoCUS (Ours) & 0.92 & 1.69 & 0.53 & 22.1 & 34.5 & 0.58 \\ 
\bottomrule
\end{tabularx}
\end{table}


\subsection{Ablation study}%
\label{sec:ablation}

We also evaluated the relative impact of different design decisions in our method, and assessed its robustness to different hyper-parameter choices.
The results from the preceding sections used the optimal combination under the constraint of similar memory consumption found in this study.
We refer the interested reader to the supplemental material for detailed results.

\section{Conclusion}
\label{sec:conclusion}

We have proposed a method for learning multi-scale view-invariant features from posed images by optimizing a novel retrieval-based objective: Vectorized-Smooth-AP. This objective modulates the DINO \cite{caron2021emerging} ViT features towards 3D-consistency and adaptively selects highly-distinguishable landmarks. Moreover, we select the retrieval set in such a way to encourage the model to balance retrieval (distinctiveness) with reusability (generalisability), through the introduction of a ``don't-care'' region beyond a certain spatial extent.

We demonstrate compelling performance when using these features for several downstream tasks, including place recognition and retrieval, semantic and instance segmentation with re-identification, and relative pose estimation, demonstrating the utility of our learned features.
This result reinforces the strong semantic properties of self-supervised image features and shows how aggregating information in 3D, via the ranking loss function and camera pose supervision, can improve their effectiveness, especially for 3D-aware tasks.
Nonetheless, strategies for removing the weak camera pose supervision warrant investigation, since a fully self-supervised approach would facilitate access to greater quantities of data. Depending on the environment, Structure-from-Motion \cite{schonberger2016structure} or SparsePose \cite{sinha2022sparsepose} may be able to alleviate this requirement, making it possible to train on larger-scale video data.

\paragraph{Ethics and attribution.}
We use the Matterport3D dataset \cite{chang2017matterport3d} in a manner compatible with their terms and the end user license agreement, available at this URL: \url{https://kaldir.vc.in.tum.de/matterport/MP_TOS.pdf}.
The dataset may accidentally contain personal data, but there is no extraction of personal or biometric information in this research.

\paragraph{Acknowledgements.} We are grateful for funding from EPSRC AIMS CDT EP/S024050/1 (D.K.), Continental AG (D.C.), and the Royal Academy of Engineering (RF/201819/18/163, J.H.).

{\small
\bibliographystyle{ieee_fullname}
\bibliography{shortstrings,vgg_local,vgg_other,refs}
}

\end{document}